\documentclass[10pt,twocolumn,letterpaper]{article}

\usepackage{dicta}
\usepackage{times}
\usepackage{epsfig}
\usepackage{graphicx}
\usepackage{amsmath}
\usepackage{amssymb}

\usepackage{booktabs}
\usepackage{adjustbox}
\usepackage{multirow}
\usepackage{makecell}
\usepackage{amssymb}
\usepackage{pifont}
\usepackage{xcolor}
\usepackage[ruled,vlined]{algorithm2e}

\usepackage[pagebackref=true,breaklinks=true,letterpaper=true,colorlinks,bookmarks=false]{hyperref}

\dictafinalcopy 

\def\dictaPaperID{73} 

\ifdictafinal\fi
\begin{document}

\title{Edge-Aware Thermal Infrared UAV Swarm Tracking}

\author{Yu-Hsi Chen\\
The University of Melbourne\\
Parkville, Australia\\
{\tt\small yuhsi@student.unimelb.edu.au}
}

\maketitle

\begin{abstract}
Thermal infrared (TIR) imaging is essential for UAV swarm operations in visually degraded environments. However, tracking tiny UAVs remains challenging due to limited appearance cues, frequent occlusions, and rapid maneuvers. Despite significant progress driven by benchmarks such as the Anti-UAV challenge, existing methods primarily prioritize accuracy while overlooking the computational constraints of real-time edge deployment. The standard Kalman Filter (KF) offers the efficiency required for edge devices, yet its constant-velocity assumption often breaks down under highly dynamic UAV motion and thermal sensor jitter. More sophisticated nonlinear estimators can improve robustness but often introduce additional computational costs. To address this gap, we propose an edge-aware online tracking pipeline centered on the Adaptive Kinematic Kalman Filter (AKKF), which augments the linear KF with state-dependent kinematic modeling while preserving real-time efficiency. Combined with transient false-positive suppression and kinematics-driven predictive coasting, the presented pipeline improves trajectory continuity under challenging TIR conditions. Experiments on the Beyond Strong Baseline (BSB) benchmark provide a starting point for edge-aware UAV tracking by jointly evaluating tracking performance and computational efficiency, offering insights toward future real-time deployment.
\end{abstract}

\section{Introduction}
\label{sec:introduction}
Cooperative UAV swarms have become increasingly important in applications such as surveillance, search and rescue, and environmental monitoring~\cite{alzahrani2020uav,abdelkader2021aerial,li2023vg}. TIR imaging has emerged as a key sensing modality for these applications, providing reliable perception under nighttime and adverse weather conditions where RGB cameras become ineffective~\cite{berg2016detection,vollmer2020infrared,he2021infrared}. However, enabling real-time multi-object tracking (MOT) on resource-constrained edge devices remains challenging, requiring a balance between tracking accuracy and computational efficiency.

\begin{figure}[t]
    \begin{center}
    \includegraphics[width=1\linewidth]{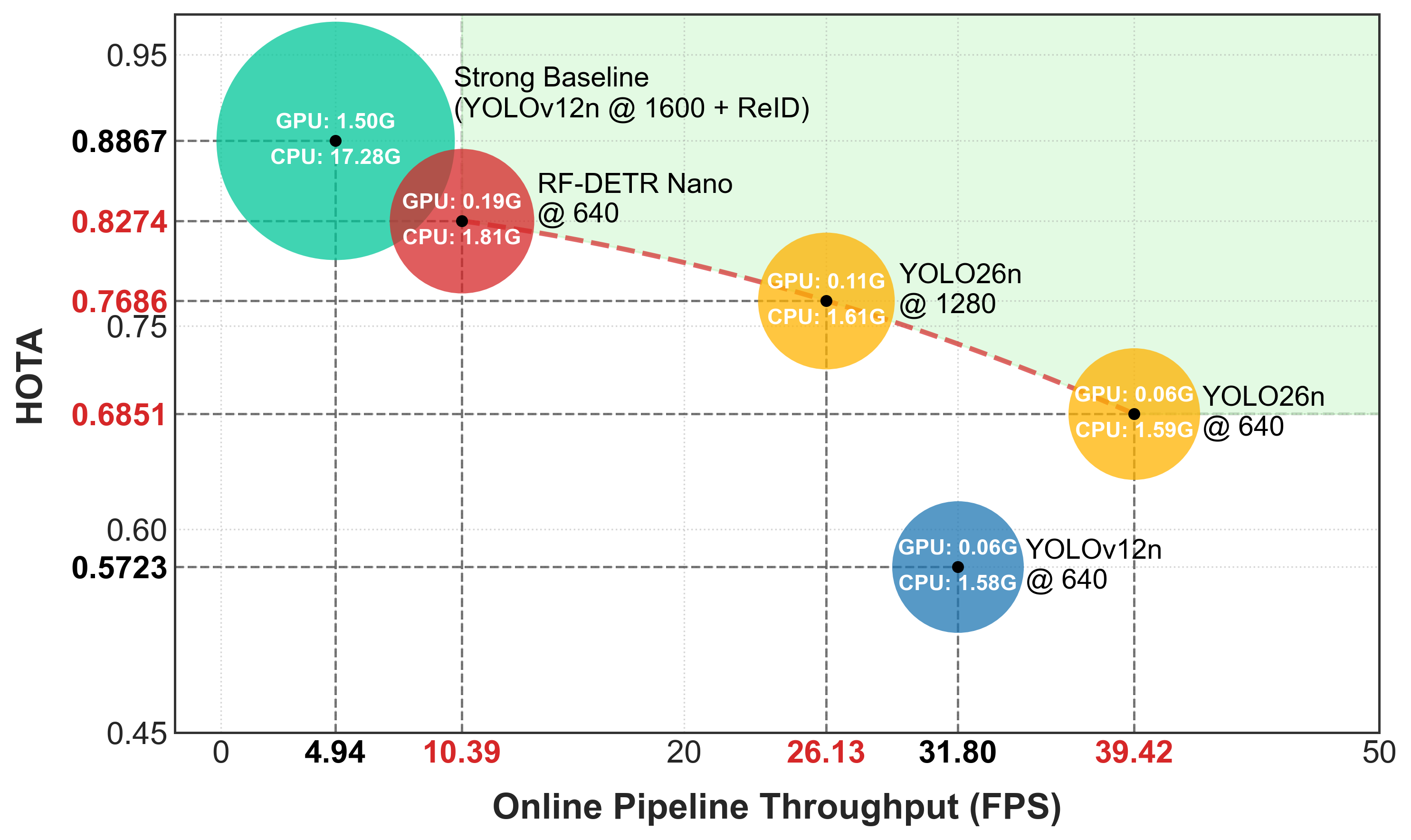}
    \end{center}
    \caption{FPS-HOTA comparison of online tracking pipelines with the proposed AKKF on the BSB leaderboard using an Intel Core i7-12650H CPU, NVIDIA GeForce RTX 4050 Laptop GPU with 6GB memory, and 24GB RAM. The x-axis denotes online throughput (FPS), and the y-axis denotes HOTA. Bubble size represents computational resource usage. The red dashed line indicates the edge-aware threshold, and the light green region indicates the preferred operating regime for edge deployment.}
    \label{fig:bubble}
\end{figure}

Several MOT pipelines follow the tracking-by-detection (TBD) paradigm, combining object detectors with data association methods such as ByteTrack~\cite{zhang2022bytetrack} and BoT-SORT~\cite{aharon2022bot}. Recent detector-tracker frameworks, such as YOLOv12 combined with BoT-SORT~\cite{chen2025strong}, have achieved competitive performance in the Anti-UAV challenge~\cite{antiuav2025challenge}. Although the KF remains the dominant motion model for online tracking due to its simplicity and efficiency, its fixed kinematic assumptions limit robustness under dynamic UAV motion. To address this limitation, we revisit the conventional KF and introduce adaptive kinematic modeling that preserves real-time efficiency while moving toward a more favorable balance between tracking performance and computational efficiency, as illustrated in Fig.~\ref{fig:bubble}. Our contributions are summarized as follows:
\begin{itemize}
    \item \textbf{Adaptive Kinematic Kalman Filter:} We augment the standard linear KF with state-dependent kinematic modeling for robust, real-time UAV swarm tracking.
    \item \textbf{Edge-aware online tracking pipeline:} We integrate TFPS and kinematics-driven predictive coasting for reliable and continuous TIR UAV tracking.
    \item \textbf{Evaluation on the BSB benchmark:} Extensive experiments characterize the accuracy-efficiency trade-off of tracking pipelines, providing a practical reference for edge-aware UAV tracking.
\end{itemize}

\section{Related Work}
\label{sec:related_work}

\subsection{UAV Datasets and Benchmarks}
\label{ssec:uav_datasets}
UAVs have been widely adopted in computer vision applications, ranging from air-to-ground target detection~\cite{bondi2020birdsai,suo2023hit} to cooperative multi-agent perception~\cite{feng2024u2udata,feng2026u2udata}. Correspondingly, UAV perception datasets have been developed across diverse sensing modalities and viewpoints, including detection and segmentation from ground-to-air and air-to-air perspectives~\cite{chen2024uavdb,guo2025yolomg}, UAV swarm monitoring~\cite{wang2022uavswarm,wang2022lightweight,dong2025securing}, and TIR UAV detection and tracking~\cite{jiang2021anti,huang2023anti}.

Despite the growing availability of UAV datasets, standardized benchmarks for multi-UAV tracking remain limited. Existing studies, such as~\cite{zhu2025gm}, often rely on custom datasets, leading to diverse data characteristics and evaluation protocols. The well-established Anti-UAV benchmarks~\cite{jiang2021anti,huang2023anti,antiuav2025challenge} provide formal evaluation platforms for TIR UAV tracking; however, their evaluation availability is limited, and they do not explicitly consider computational efficiency for edge deployment. Building upon the publicly released Anti-UAV data~\cite{antiuav2025challenge}, the BSB benchmark~\cite{chen2025strong} provides a continuously maintained multi-UAV tracking evaluation platform\footnotemark[1] with carefully train/test splits, data characteristics summarized in Tab.~\ref{tab:dataset_stats}, and HOTA-based evaluation protocols~\cite{luiten2021hota}, enabling reproducible comparison of tracking approaches under challenging TIR UAV scenarios.
\footnotetext[1]{\href{https://www.codabench.org/competitions/16223/}{BSB benchmark website}}

\begin{table}[t]
\begin{center}
\begin{adjustbox}{max width=\linewidth}
\renewcommand{\arraystretch}{1.1}
\setlength{\tabcolsep}{8pt}
\begin{tabular}{lcc}
\toprule
\textbf{Characteristic} & \textbf{Training Data}\footnotemark[2] & \textbf{Testing Data}\footnotemark[3]\(^{,}\)\footnotemark[4] \\
\midrule
Number of Sequences & 102 & 98 \\
Number of Frames & 77,293 & 74,537 \\
Resolutions & $640 \times 512$ & $640 \times 512$ \\
Total Bounding Boxes & 1,633,110 & 1,950 \\
Width Range (px) & [1.43, 72.50] & [3.60, 60.68] \\
Width Mean $\pm$ Std (px) & 10.68 $\pm$ 5.80 & 10.49 $\pm$ 5.27 \\
Height Range (px) & [0.88, 52.84] & [3.17, 45.12] \\
Height Mean $\pm$ Std (px) & 8.83 $\pm$ 4.43 & 9.49 $\pm$ 4.91 \\
Area Range (px$^2$) & [2.36, 3128.69] & [11.92, 2126.08] \\
Area Mean $\pm$ Std (px$^2$) & 116.90 $\pm$ 158.98 & 122.73 $\pm$ 189.04 \\
\bottomrule
\end{tabular}
\end{adjustbox}
\end{center}
\caption{Summary of dataset characteristics from the BSB benchmark. The benchmark is curated from 200 public videos with a sequence-wise split to prevent information leakage and maintain consistent scenario coverage between training and testing sets. The testing set provides only first-frame bounding box annotations for tracker initialization and evaluation.}
\label{tab:dataset_stats}
\end{table}
\footnotetext[2]{\href{https://doi.org/10.5281/zenodo.15853476}{BSB training dataset}}
\footnotetext[3]{\href{https://doi.org/10.5281/zenodo.16299533}{BSB test dataset}}
\footnotetext[4]{\href{https://doi.org/10.5281/zenodo.16601508}{Corrected test sequence BB2P\_02}}

\subsection{Real-Time Multi-Object Tracking}
\label{ssec:real_time_mot}
Recent studies have explored data-driven state estimators to model complex motion dynamics~\cite{revach2021kalmannet,revach2022kalmannet,choi2023split,buchnik2023latent,ni2024adaptive,buchnik2024gsp,chen2025maml,dahan2025bayesian}. Meanwhile, efficient motion-centric association strategies have demonstrated competitive performance for real-time tracking. Motion-based trackers, such as ByteTrack~\cite{zhang2022bytetrack} and OCSORT~\cite{cao2023observation}, achieve effective association by leveraging bounding-box kinematics and observation history. Hybrid trackers, including StrongSORT~\cite{du2023strongsort} and BoT-SORT~\cite{aharon2022bot}, further incorporate ReID modules and camera motion compensation (CMC) to enhance association performance while maintaining real-time efficiency.

At the core of most motion-based trackers lies the discrete-time KF~\cite{kalman1960new,bar2001estimation}, which provides an efficient recursive framework for state estimation under linear dynamics. Given a target state $\boldsymbol{x}_k \in \mathbb{R}^{n}$ and noisy observation $\boldsymbol{z}_k \in \mathbb{R}^{m}$ at time step $k \in \mathbb{N}^{+}$, the system is formulated as:
\begin{align}
    \boldsymbol{x}_k &= \boldsymbol{F}_k\boldsymbol{x}_{k-1}+\boldsymbol{w}_{k},\\
    \boldsymbol{z}_k &= \boldsymbol{H}_k\boldsymbol{x}_k+\boldsymbol{v}_k,
\end{align}
where $\boldsymbol{F}_k$ and $\boldsymbol{H}_k$ denote the state transition and observation matrices, respectively. Without external control inputs in bounding-box MOT, the dynamics are governed by kinematic evolution with Gaussian process and measurement noises, $\boldsymbol{w}_k\sim\mathcal{N}(\boldsymbol{0},\boldsymbol{Q}_k)$ and $\boldsymbol{v}_k\sim\mathcal{N}(\boldsymbol{0},\boldsymbol{R}_k)$. The prior estimate is propagated through the prediction step:
\begin{align}
    \hat{\boldsymbol{x}}_{k|k-1} &= \boldsymbol{F}_k\hat{\boldsymbol{x}}_{k-1|k-1}, \\
    \boldsymbol{P}_{k|k-1} &= \boldsymbol{F}_k\boldsymbol{P}_{k-1|k-1}\boldsymbol{F}_k^\top+\boldsymbol{Q}_k.
\end{align}
The Kalman gain $\boldsymbol{K}_k$ is then computed in the update step to fuse the prior prediction with the incoming measurement:
\begin{align}
    \boldsymbol{K}_k &= \boldsymbol{P}_{k|k-1}\boldsymbol{H}_k^\top
    (\boldsymbol{H}_k\boldsymbol{P}_{k|k-1}\boldsymbol{H}_k^\top+\boldsymbol{R}_k)^{-1}, \\
    \hat{\boldsymbol{x}}_{k|k} &= \hat{\boldsymbol{x}}_{k|k-1}
    +\boldsymbol{K}_k(\boldsymbol{z}_k-\boldsymbol{H}_k\hat{\boldsymbol{x}}_{k|k-1}), \\
    \boldsymbol{P}_{k|k} &= (\boldsymbol{I} - \boldsymbol{K}_k\boldsymbol{H}_k)\boldsymbol{P}_{k|k-1},
\end{align}
where $\boldsymbol{K}_k$ adaptively balances motion prediction and detector measurements according to their uncertainties.

Nonlinear KF variants, such as the Extended Kalman Filter (EKF)~\cite{anderson2005optimal} and Unscented Kalman Filter (UKF)~\cite{julier1997new}, provide greater modeling flexibility for complex dynamics but are rarely adopted in bounding-box MOT due to their increased computational cost for multiple targets. Consequently, SOTA trackers, including recent pipelines~\cite{tu2026tracking,guo2026enbot}, continue to rely on the standard KF with linear constant-velocity assumptions and predefined noise models~\cite{bewley2016simple}, including TIR UAV tracking systems~\cite{qin2025pptracker,wang2025dist,chen2025strong}.

\section{Methodology}
\label{sec:methodology}

\subsection{Adaptive Kinematic Kalman Filter}
\label{ssec:akkf}
We extend the linear KF with state-dependent kinematic modeling while preserving real-time efficiency. Following BoT-SORT~\cite{aharon2022bot}, AKKF adaptively updates the transition and measurement uncertainty:
\begin{equation}
    \boldsymbol{F}_k = \boldsymbol{F}_k(\boldsymbol{x}_{k-1}), \quad
    \boldsymbol{R}_k = \boldsymbol{R}_k(\boldsymbol{z}_k).
\end{equation}
The state and measurement vectors are defined as:
\begin{align}
    \boldsymbol{x}_k &=
    [x_c(k), y_c(k), w(k), h(k),\\
    & \hspace*{0.58cm} v_x(k), v_y(k), v_w(k), v_h(k)]^\top, \\
    \boldsymbol{z}_k &=
    [z_{x_c}(k), z_{y_c}(k), z_w(k), z_h(k)]^\top,
\end{align}
where $(x_c,y_c,w,h)$ denote the bounding-box center and scale, $(v_x,v_y,v_w,v_h)$ represent the corresponding velocities, and $\boldsymbol{z}_k$ represents the detector measurement at frame $k$. The frame interval is set to $\Delta t=1$.

\subsubsection{Area-Adaptive Motion Damping}
\label{sssec:area_adaptive_motion_damping}
The standard KF assumes constant positional velocity by setting $\boldsymbol{F}_{5,5}$ and $\boldsymbol{F}_{6,6}$ to one. During prolonged occlusions, this assumption leads to trajectory drift as the last estimated velocity is continuously propagated. To address this issue, we introduce Area-Adaptive Motion Damping (AAMD), which dynamically dampens positional velocity based on target scale. Inspired by aerodynamic damping models~\cite{bar2001estimation}, the damping factor is defined as:
\begin{equation}
    D_{xy} = \alpha
    \left(\frac{A_{box}}{A_{box}+\beta}\right),
\end{equation}
where $A_{box}=\max(1,w\times h)$ is the bounding-box area, $\alpha$ controls the maximum damping strength, and $\beta$ determines the saturation behavior. The velocity retention terms are replaced by $(1-D_{xy})$, allowing velocity decay during occlusions while maintaining trajectory continuity.

\subsubsection{Perspective-Aware Velocity Braking}
\label{sssec:perspective_aware_velocity_braking}
The standard KF assumes constant scale velocity by setting $\boldsymbol{F}_{7,7}$ and $\boldsymbol{F}_{8,8}$ to one. When a UAV recedes from the camera, negative scale velocities with $v_w<0$ and $v_h<0$ may lead to scale collapse if detections become unavailable. To mitigate this issue, we introduce Perspective-Aware Velocity Braking (PAVB), which dynamically suppresses scale velocity based on the target scale:
\begin{equation}
    D_w = \gamma\frac{1}{\max(w^{(k-1)},\epsilon)}, \quad
    D_h = \gamma\frac{1}{\max(h^{(k-1)},\epsilon)},
\end{equation}
where $\epsilon$ prevents numerical instability and $\gamma$ controls the braking strength. The coefficients modify the scale velocity retention terms to reduce excessive scale variation.

\subsubsection{Dynamic Measurement Noise Estimation}
\label{sssec:dynamic_measurement_noise_estimation}
The standard BoT-SORT measurement model assumes a scale-dependent measurement covariance $\boldsymbol{R}_k$. However, TIR clutter can introduce geometrically inconsistent detections beyond scale-based uncertainty. To address this issue, we propose Dynamic Measurement Noise Estimation (DMNE), which adjusts measurement uncertainty according to detection consistency. Given the measured and predicted aspect ratios, $a_{meas}=w_{meas}/h_{meas}$ and $a_{pred}=w_{pred}/h_{pred}$, respectively, we define an anomaly factor:
\begin{equation}
    \rho =
    1+\kappa
    \left(\frac{|a_{meas}-a_{pred}|}{a_{pred}}\right)^2,
\end{equation}
where $\kappa$ controls the sensitivity to aspect-ratio deviations. The measurement covariance is then updated as:
\begin{equation}
    \boldsymbol{R}_k(\boldsymbol{z}_k)=\rho\boldsymbol{R}_k.
\end{equation}
Large geometric inconsistencies increase $\rho$, which enlarges $\boldsymbol{R}_k$ and reduces the Kalman gain contribution from unreliable detections, allowing the prediction to dominate.

\subsection{Online Tracking Pipeline}
\label{sssec:online_tracking_pipeline}
Beyond the proposed AKKF, we further enhance the online tracking pipeline with two practical components tailored for TIR UAV tracking, namely Transient False Positive Suppression (TFPS) to improve track reliability and kinematics-guided predictive coasting to maintain trajectory continuity under temporary observation gaps.

\subsubsection{Transient False Positive Suppression}
\label{sssec:baseline_correction}
The standard BoT-SORT implementation outputs newly initialized tracklets before temporal confirmation, allowing one-frame false positives to appear in the final tracking results. This setup is particularly problematic in TIR imagery, where thermal clutter frequently produces transient false detections. To address this issue, we restore temporal track confirmation, referred to as TFPS, by outputting only confirmed tracklets.
As shown in Fig.~\ref{fig:coasting}, this refinement suppresses transient false positives and improves the baseline HOTA from $0.82357$ to $0.825405$. All subsequent experiments are conducted on this updated baseline, except for the results reported in Tab.~\ref{tab:det_vs_track}.

\subsubsection{Kinematics-Guided Predictive Coasting}
\label{sssec:coasting}
\begin{figure}[t]
    \begin{center}
    \includegraphics[width=\linewidth]{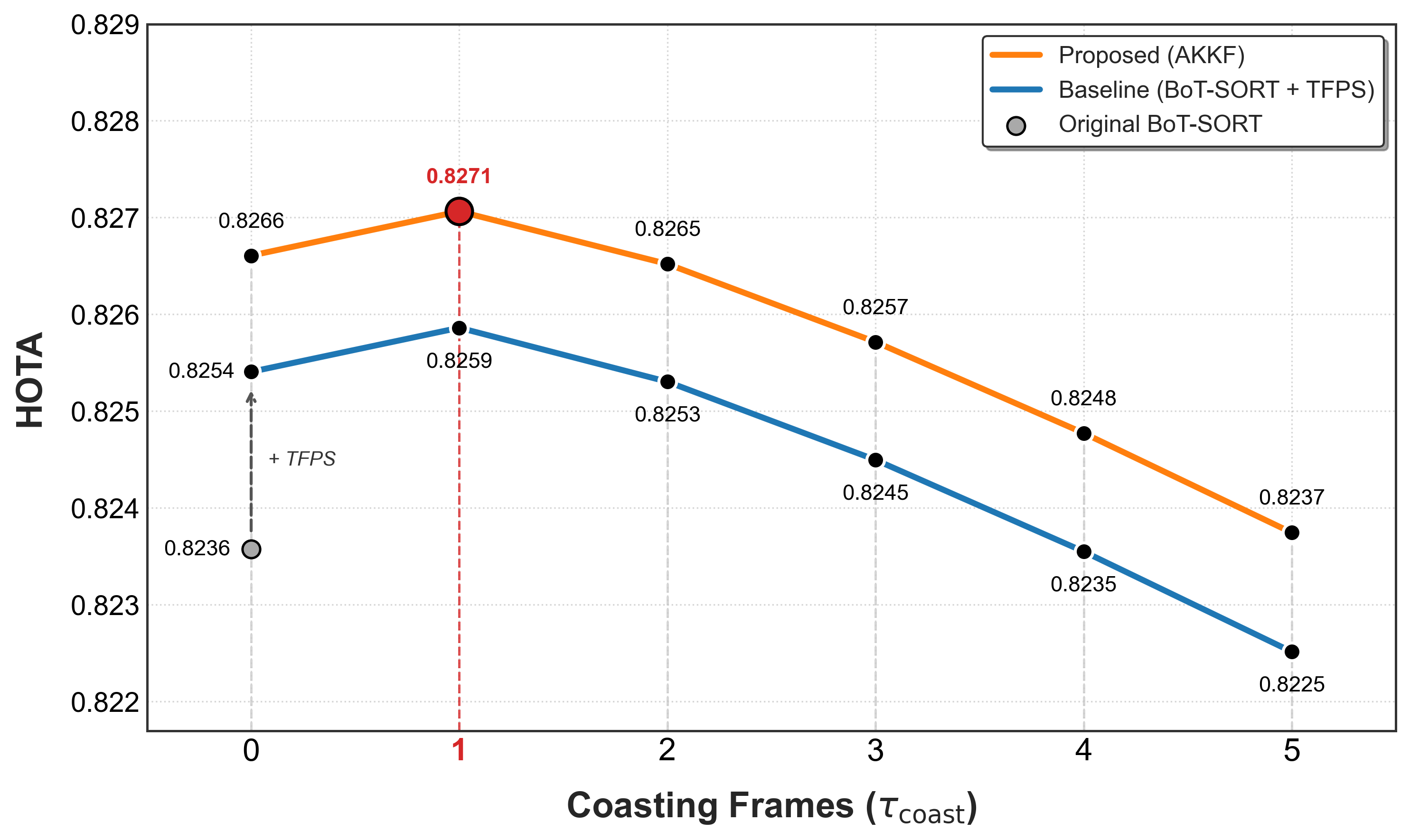}
    \end{center}
    \caption{Effect of TFPS and predictive coasting on the BSB leaderboard using the RF-DETR Nano tracking pipeline with 640 input resolution. HOTA is plotted against the number of predictive coasting frames. The vertical gray dashed line indicates the performance gain from TFPS, as discussed in Sec.~\ref{sssec:baseline_correction}. Both the standard BoT-SORT and the proposed AKKF achieve peak HOTA using a single predictive coasting frame.}
    \label{fig:coasting}
\end{figure}

Temporary missed detections could be caused by thermal sensor jitter, motion blur, or mutual occlusions, which can interrupt trajectory association in TIR UAV tracking. To improve track continuity, we incorporate a predictive coasting strategy. When no valid detection is associated with frame $k$, the measurement update is skipped, and the prediction is directly adopted as the posterior estimate:
\begin{align}
    \hat{\boldsymbol{x}}_{k|k}
    &= \hat{\boldsymbol{x}}_{k|k-1}
    = \boldsymbol{F}_k\hat{\boldsymbol{x}}_{k-1|k-1},\\
    \boldsymbol{P}_{k|k}
    &= \boldsymbol{P}_{k|k-1}
    = \boldsymbol{F}_k\boldsymbol{P}_{k-1|k-1}\boldsymbol{F}_k^\top
    + \boldsymbol{Q}_k.
\end{align}
We define $\tau_{\mathrm{coast}}$ as the number of consecutive missed associations handled by predictive coasting. As shown in Fig.~\ref{fig:coasting}, both the standard KF and the proposed AKKF achieve the best performance with a single coasting frame using the RF-DETR Nano tracking pipeline with 640 input resolution.

\section{Experiment Results}
\label{sec:experimental_results}

\subsection{Implementation Details}
\label{ssec:exp_setup}
We evaluate several advanced detectors, including YOLOv12~\cite{tian2026yolov12}, YOLO26~\cite{Jocher_Ultralytics_YOLO26_Unified_2026}, and RF-DETR~\cite{Robinson_RF-DETR_2025}, integrated with the BoT-SORT tracker~\cite{aharon2022bot} and the proposed AKKF. We primarily report HOTA metrics from the BSB leaderboard evaluation~\cite{chen2025strong} introduced in Sec.~\ref{ssec:uav_datasets}. The model configurations, including model size and input resolution, are presented in Tab.~\ref{tab:det_vs_track}. Unless otherwise specified, the batch size is set to 32, while YOLOv12s and YOLOv12m use batch sizes of 16 and 8, respectively.

All experiments are conducted on an NVIDIA H100 80GB GPU, except for Fig.~\ref{fig:bubble} and Tab.~\ref{tab:ablation}, which use an NVIDIA GeForce RTX 4050 Laptop GPU with 6GB memory to better approximate edge deployment conditions. The slight HOTA difference between the offline result of $0.8271$ in Fig.~\ref{fig:coasting} and the online result of $0.8274$ in Fig.~\ref{fig:bubble} is due to different processing pipelines: the offline pipeline truncates detection outputs to two decimal places before tracking, while the online pipeline directly uses original outputs.

\subsection{Main Results}
\label{ssec:main_results}

Tab.~\ref{tab:det_vs_track} summarizes the results, including training time, inference speed, peak GPU memory usage, detection accuracy, and tracking performance. RF-DETR Nano with 1280 resolution achieves the highest HOTA score of $0.8437$. The underlined configurations, including RF-DETR Nano with 640 resolution and YOLO26n with 1280 and 640 resolutions, are selected as edge-aware candidates and further evaluated in Fig.~\ref{fig:bubble} for resource-constrained deployment.

\begin{figure*}[t]
    \begin{center}
    \includegraphics[width=1\linewidth]{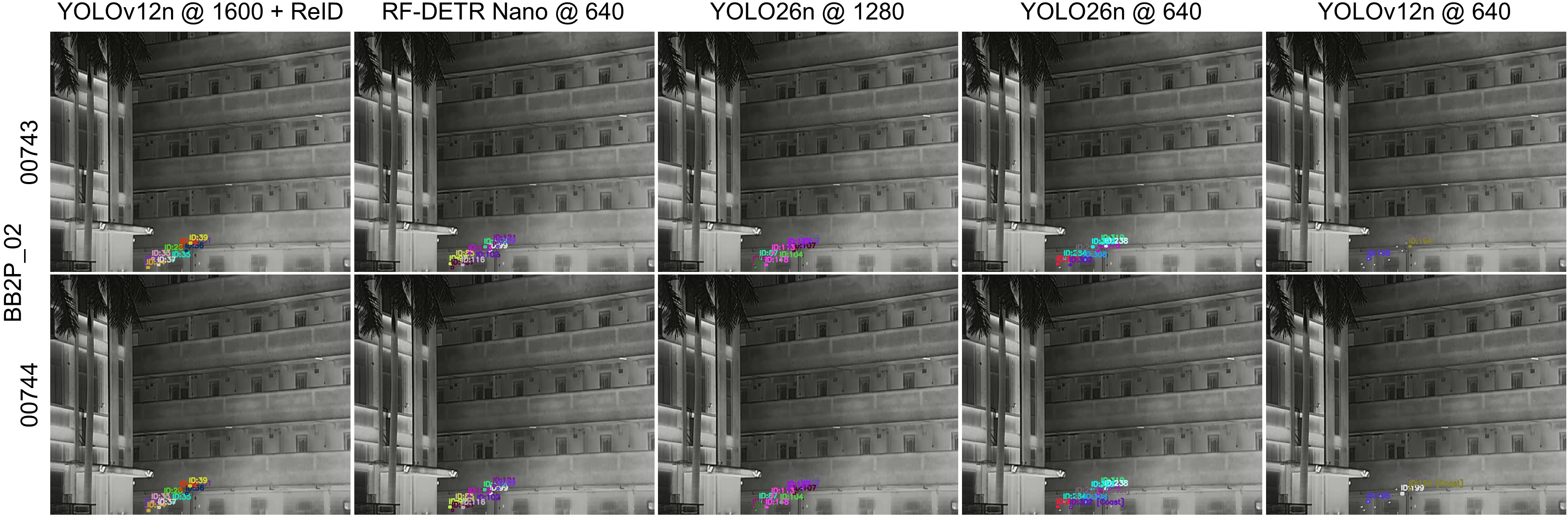}
    \end{center}
    \caption{Visualization of tracked bounding boxes from different pipelines across two consecutive frames of the BB2P\_02 sequence.}
    \label{fig:demo}
\end{figure*}

The visualization comparison in Fig.~\ref{fig:demo} illustrates the tracking behavior of different configurations. The leftmost column shows the strong baseline from~\cite{chen2025strong} with the best localization and identity preservation, while the three middle panels present the edge-aware candidates with gradually decreasing tracking quality from left to right. YOLO26n at 640 resolution achieves a superior accuracy-throughput trade-off over YOLOv12n at the same resolution, consistent with Fig.~\ref{fig:bubble}. The temporal evolution from the upper to lower rows in Fig.~\ref{fig:demo}, labeled [Coast], demonstrates effective predictive coasting during temporary observation gaps.

\begin{table*}[t]
\begin{center}
\begin{adjustbox}{max width=\linewidth, max height=0.5\textheight}
\renewcommand{\arraystretch}{1.1}
\setlength{\tabcolsep}{8pt}
\begin{tabular}{ccc|cc|cc|cc}
\toprule
\multirow{3.5}{*}{\textbf{Model}} & \multirow{3.5}{*}{\textbf{Scale}} & \multirow{3.5}{*}{\makecell{\textbf{Size} \\ \textbf{(pixels)}}} & \multicolumn{2}{c|}{\textbf{Train}} & \multicolumn{2}{c|}{\textbf{Inference}} & \multirow{3.5}{*}{\textbf{$\text{AP}^{val}_{50-95} \uparrow$}} & \multirow{3.5}{*}{\textbf{HOTA $\uparrow$}} \\
\cmidrule{4-7}
 & & & \makecell{\textbf{Peak GPU} \\ \textbf{RAM (GB)}} \(\downarrow\) & \makecell{\textbf{Total Time} \\ \textbf{(hh:mm:ss)}} \(\downarrow\) & \makecell{\textbf{Peak GPU} \\ \textbf{RAM (GB)}} \(\downarrow\) & \makecell{\textbf{Total Time} \\ \textbf{(hh:mm:ss)}} \(\downarrow\) & & \\
\midrule
\multirow{6}{*}{YOLOv12} & n & 1280 & 54.011 & 07:20:20 & 1.042 & 00:26:22 & 0.473 & 0.717086 \\
                         & s & 1280 & 54.022 & 12:42:55 & 0.958 & 00:25:47        & 0.477 & 0.715683 \\
                         & m & 1280 & 47.181 & 21:56:43 & 1.257 & 00:28:07        & 0.483 & 0.713188 \\
                         \cmidrule{2-9}
                         & n & 640  & 13.688 & 02:09:41 & 0.868 & 00:20:57 & 0.433 & 0.521526 \\
                         & s & 640  & 20.452 & 02:36:01 & 0.892 & 00:24:30        & 0.436 & 0.552467 \\
                         & m & 640  & 27.878 & 03:38:52 & 0.925 & 00:20:25        & 0.444 & 0.551958 \\
\midrule
\multirow{6}{*}{YOLO26}  & n & 1280 & 40.835 & 05:13:49 & 2.026 & 00:11:24 & 0.472 & \underline{0.764241} \\
                         & s & 1280 & 65.396 & 06:13:00 & 3.073 & 00:12:53 & 0.464 & 0.748453 \\
                         & m & 1280 & 75.624 & 08:27:27 & 5.464 & 00:15:32 & 0.465 & 0.788816 \\
                         \cmidrule{2-9}
                         & n & 640  & 10.560 & 03:02:54 & 1.050 & 00:05:38 & 0.448 & \underline{0.681414} \\
                         & s & 640  & 14.542 & 03:15:21 & 1.315 & 00:05:59 & 0.452 & 0.693846 \\
                         & m & 640  & 20.659 & 04:00:13 & 1.892 & 00:06:07 & 0.444 & 0.711008 \\
\midrule
\multirow{6}{*}{RF-DETR} & n & 1280 & 31.327 & 17:12:11 & 1.144 & 00:45:22 & 0.435 & \textbf{0.843723} \\
                         & s & 1280 & 31.931 & 18:44:13 & 1.144 & 00:44:40 & 0.437 & 0.831454 \\
                         & m & 1280 & 33.030 & 20:06:52 & 1.144 & 00:44:58 & 0.437 & 0.835436 \\
                         \cmidrule{2-9}
                         & n & 640  & 13.349 & 11:31:36 & 0.906 & 00:31:16 & 0.422 & \underline{0.823570} \\
                         & s & 640  & 18.747 & 13:19:50 & 0.906 & 00:32:50 & 0.427 & 0.815145 \\
                         & m & 640  & 17.681 & 14:50:46 & 0.906 & 00:34:51 & 0.415 & 0.809826 \\
\bottomrule
\end{tabular}
\end{adjustbox}
\end{center}
\caption{Comprehensive comparison of detection architectures on the BSB leaderboard. All measurements are obtained using an NVIDIA H100 80GB GPU. Peak training memory, inference time, detection precision, and tracking performance with the original BoT-SORT tracker are reported. The best-performing detector is highlighted in \textbf{bold}, and edge-aware candidates are \underline{underlined}.}
\label{tab:det_vs_track}
\end{table*}

\subsection{Ablation Study and Discussion}
\label{ssec:ablation_studies_and_discussion}

\begin{figure}[t]
    \begin{center}
    \includegraphics[width=\linewidth]{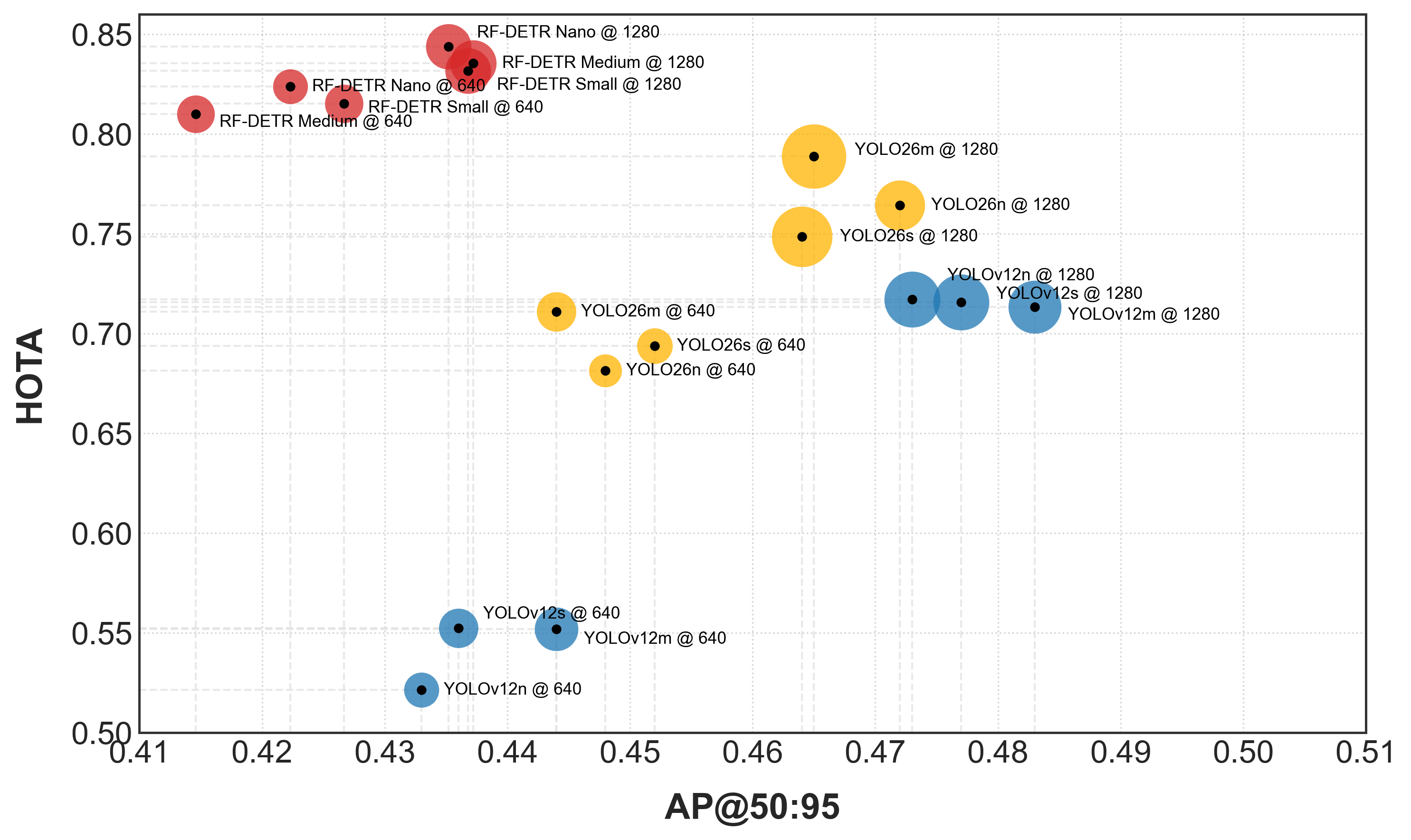}
    \end{center}
    \caption{AP@50:95-HOTA comparison of tracking pipelines on the BSB leaderboard. The x-axis denotes detection performance measured by AP@50:95, and the y-axis denotes tracking performance measured by HOTA. The results reveal that higher detection accuracy does not necessarily lead to better tracking performance, exposing the detection-tracking paradox.}
    \label{fig:paradox}
\end{figure}

Tab.~\ref{tab:ablation} summarizes the contribution of each AKKF component. The full AKKF achieves the best overall tracking performance, improving all metrics over the baseline while reducing FPS from $132.03$ to $94.92$. 
Beyond the ablation results, Fig.~\ref{fig:paradox} shows that higher detection accuracy does not necessarily translate into better tracking performance, revealing the detection-tracking paradox also observed in~\cite{chen2025strong}. For tiny-object tracking, increasing input resolution consistently improves performance, aligning with previous findings~\cite{chen2023one,chen2025strong}. Since UAV targets are already extremely small, further reducing resolution may severely degrade tracking quality. Therefore, future improvements should focus on computational optimization, such as model pruning~\cite{fang2023depgraph,chen2024optimizing,fang2024isomorphic}. Additionally, under edge constraints with limited appearance cues in TIR imagery, leveraging temporal information through motion modeling offers a promising direction.

\begin{table}[t]
\begin{center}
\begin{adjustbox}{max width=\linewidth, max height=0.5\textheight}
\renewcommand{\arraystretch}{1.1}
\setlength{\tabcolsep}{8pt}
\begin{tabular}{ccccccc} 
\toprule
\multicolumn{3}{c}{\textbf{AKKF Components}} & \multicolumn{3}{c}{\textbf{Tracking Accuracy}} & \textbf{Efficiency} \\
\cmidrule(lr){1-3} \cmidrule(lr){4-6} \cmidrule(lr){7-7}
\textbf{AAMD} & \textbf{PAVB} & \textbf{DMNE} & \textbf{HOTA $\uparrow$} & \textbf{MOTA $\uparrow$} & \textbf{IDF1 $\uparrow$} & \textbf{FPS $\uparrow$} \\
\midrule
           &            &            & 0.826116 & 0.636194 & 0.687265 & \textbf{132.03} \\
\checkmark &            &            & 0.826154 & 0.636183 & 0.687287 & 129.71 \\
           & \checkmark &            & 0.826196 & 0.636275 & 0.687395 & 129.89 \\
           &            & \checkmark & 0.826441 & 0.636180 & 0.687744 & 126.05 \\
\checkmark & \checkmark &            & 0.826240 & \textbf{0.636296} & 0.687429 & 127.91 \\
\checkmark &            & \checkmark & 0.826462 & 0.636185 & 0.687766 & 123.56 \\
           & \checkmark & \checkmark & 0.826536 & 0.636253 & 0.687949 & 126.05 \\
\checkmark & \checkmark & \checkmark & \textbf{0.826606} & 0.636284 & \textbf{0.688019} & \hspace*{1.9mm}94.92  \\
\bottomrule
\end{tabular}
\end{adjustbox}
\end{center}
\caption{Ablation study of the proposed AKKF components, including AAMD, PAVB, and DMNE, using an Intel Core i7-12650H CPU, NVIDIA GeForce RTX 4050 Laptop GPU with 6GB memory, and 24GB RAM. The top-row baseline denotes the standard BoT-SORT with constant-velocity kinematics, TFPS, and no predictive coasting, while AKKF introduces adaptive kinematic modeling through the proposed components. Best results for each tracking metric are highlighted in \textbf{bold}.}
\label{tab:ablation}
\end{table}

\section{Conclusion}
\label{sec:conclusion}
In this paper, we present an edge-aware tracking pipeline for TIR UAV swarms that jointly considers tracking robustness and computational efficiency. We propose the AKKF, which augments the standard KF framework with adaptive kinematic priors through state-dependent motion damping, velocity constraints, and measurement uncertainty estimation. Combined with TFPS and predictive coasting, the proposed pipeline enables robust TIR UAV tracking under temporary observation gaps. Extensive experiments on the BSB benchmark validate its effectiveness and highlight the detection-tracking paradox in TIR UAV tracking. Future work will explore model compression, temporal information utilization, and validation on dedicated edge devices.

\section{Acknowledgments}
\label{sec:acknowledgements}
We thank the Spartan HPC system~\cite{meade2017spartan} at The University of Melbourne for providing the computational resources that accelerated model training and supported this research.

{\small
\bibliographystyle{ieee}
\bibliography{egbib}

@article{alzahrani2020uav,
  title={UAV assistance paradigm: State-of-the-art in applications and challenges},
  author={Alzahrani, Bander and Oubbati, Omar Sami and Barnawi, Ahmed and Atiquzzaman, Mohammed and Alghazzawi, Daniyal},
  journal={Journal of Network and Computer Applications},
  volume={166},
  pages={102706},
  year={2020},
  publisher={Elsevier}
}

@article{abdelkader2021aerial,
  title={Aerial swarms: Recent applications and challenges},
  author={Abdelkader, Mohamed and G{\"u}ler, Samet and Jaleel, Hassan and Shamma, Jeff S},
  journal={Current robotics reports},
  volume={2},
  number={3},
  pages={309--320},
  year={2021},
  publisher={Springer}
}

@article{li2023vg,
  title={VG-swarm: A vision-based gene regulation network for UAVs swarm behavior emergence},
  author={Li, Huanlin and Cai, Yuwei and Hong, Juncao and Xu, Peng and Cheng, Hui and Zhu, Xiaomin and Hu, Bingliang and Hao, Zhifeng and Fan, Zhun},
  journal={IEEE Robotics and Automation Letters},
  volume={8},
  number={3},
  pages={1175--1182},
  year={2023},
  publisher={IEEE}
}

@book{berg2016detection,
  title={Detection and tracking in thermal infrared imagery},
  author={Berg, Amanda},
  year={2016},
  publisher={Linkopings Universitet (Sweden)}
}

@incollection{vollmer2020infrared,
  title={Infrared thermal imaging},
  author={Vollmer, Michael},
  booktitle={Computer vision: A reference guide},
  pages={1--4},
  year={2020},
  publisher={Springer}
}

@article{he2021infrared,
  title={Infrared machine vision and infrared thermography with deep learning: A review},
  author={He, Yunze and Deng, Baoyuan and Wang, Hongjin and Cheng, Liang and Zhou, Ke and Cai, Siyuan and Ciampa, Francesco},
  journal={Infrared physics \& technology},
  volume={116},
  pages={103754},
  year={2021},
  publisher={Elsevier}
}

@misc{antiuav2025challenge,
  author       = {{Anti-UAV Challenge Organizers}},
  title        = {The 4th Anti-UAV Workshop and Challenge ({CVPR} 2025)},
  howpublished = {\url{https://anti-uav.github.io/}},
  year         = {2025}
}

@inproceedings{bondi2020birdsai,
  title={BIRDSAI: A dataset for detection and tracking in aerial thermal infrared videos},
  author={Bondi, Elizabeth and Jain, Raghav and Aggrawal, Palash and Anand, Saket and Hannaford, Robert and Kapoor, Ashish and Piavis, Jim and Shah, Shital and Joppa, Lucas and Dilkina, Bistra and others},
  booktitle={Proceedings of the IEEE/CVF Winter conference on applications of computer vision},
  pages={1747--1756},
  year={2020}
}

@article{suo2023hit,
  title={HIT-UAV: A high-altitude infrared thermal dataset for Unmanned Aerial Vehicle-based object detection},
  author={Suo, Jiashun and Wang, Tianyi and Zhang, Xingzhou and Chen, Haiyang and Zhou, Wei and Shi, Weisong},
  journal={Scientific Data},
  volume={10},
  number={1},
  pages={227},
  year={2023},
  publisher={Nature Publishing Group UK London}
}

@inproceedings{feng2024u2udata,
  title={U2udata: A large-scale cooperative perception dataset for swarm uavs autonomous flight},
  author={Feng, Tongtong and Wang, Xin and Han, Feilin and Zhang, Leping and Zhu, Wenwu},
  booktitle={Proceedings of the 32nd ACM International Conference on Multimedia},
  pages={7600--7608},
  year={2024}
}

@inproceedings{feng2026u2udata,
  title={U2UData+: A scalable swarm UAVs autonomous flight dataset for embodied long-horizon tasks},
  author={Feng, Tongtong and Wang, Xin and Han, Feilin and Zhang, Leping and Zhu, Wenwu},
  booktitle={Proceedings of the AAAI Conference on Artificial Intelligence},
  volume={40},
  number={3},
  pages={1792--1800},
  year={2026}
}

@article{chen2024uavdb,
  title={UAVDB: Point-Guided Masks for UAV Detection and Segmentation},
  author={Chen, Yu-Hsi},
  journal={arXiv preprint arXiv:2409.06490},
  year={2024}
}

@article{guo2025yolomg,
  title={YOLOMG: Vision-based Drone-to-Drone Detection with Appearance and Pixel-Level Motion Fusion},
  author={Guo, Hanqing and Lin, Xiuxiu and Zhao, Shiyu},
  journal={arXiv preprint arXiv:2503.07115},
  year={2025}
}

@article{wang2022uavswarm,
  title={UAVSwarm dataset: An unmanned aerial vehicle swarm dataset for multiple object tracking},
  author={Wang, Chuanyun and Su, Yang and Wang, Jingjing and Wang, Tian and Gao, Qian},
  journal={Remote Sensing},
  volume={14},
  number={11},
  pages={2601},
  year={2022},
  publisher={MDPI}
}

@article{wang2022lightweight,
  title={A lightweight UAV swarm detection method integrated attention mechanism},
  author={Wang, Chuanyun and Meng, Linlin and Gao, Qian and Wang, Jingjing and Wang, Tian and Liu, Xiaona and Du, Furui and Wang, Linlin and Wang, Ershen},
  journal={Drones},
  volume={7},
  number={1},
  pages={13},
  year={2022},
  publisher={MDPI}
}

@inproceedings{dong2025securing,
  title={Securing the skies: A comprehensive survey on anti-uav methods, benchmarking, and future directions},
  author={Dong, Yifei and Wu, Fengyi and Zhang, Sanjian and Chen, Guangyu and Hu, Yuzhi and Yano, Masumi and Sun, Jingdong and Huang, Siyu and Liu, Feng and Dai, Qi and others},
  booktitle={Proceedings of the Computer Vision and Pattern Recognition Conference},
  pages={6659--6673},
  year={2025}
}

@article{zhu2025gm,
  title={GM-DETR: Infrared Detection of Small UAV Swarm Targets Based on Detection Transformer},
  author={Zhu, Chenhao and Xie, Xueli and Xi, Jianxiang and Yang, Xiaogang},
  journal={Remote Sensing},
  volume={17},
  number={19},
  pages={3379},
  year={2025},
  publisher={MDPI}
}

@article{jiang2021anti,
  title={Anti-UAV: A large-scale benchmark for vision-based UAV tracking},
  author={Jiang, Nan and Wang, Kuiran and Peng, Xiaoke and Yu, Xuehui and Wang, Qiang and Xing, Junliang and Li, Guorong and Guo, Guodong and Ye, Qixiang and Jiao, Jianbin and others},
  journal={IEEE Transactions on Multimedia},
  volume={25},
  pages={486--500},
  year={2021},
  publisher={IEEE}
}

@article{huang2023anti,
  title={Anti-UAV410: A thermal infrared benchmark and customized scheme for tracking drones in the wild},
  author={Huang, Bo and Li, Jianan and Chen, Junjie and Wang, Gang and Zhao, Jian and Xu, Tingfa},
  journal={IEEE Transactions on Pattern Analysis and Machine Intelligence},
  volume={46},
  number={5},
  pages={2852--2865},
  year={2023},
  publisher={IEEE}
}

@inproceedings{chen2025strong,
  title={Strong baseline: multi-UAV tracking via YOLOv12 with bot-sort-reid},
  author={Chen, Yu-Hsi},
  booktitle={Proceedings of the Computer Vision and Pattern Recognition Conference},
  pages={6573--6582},
  year={2025}
}

@article{luiten2021hota,
  title={Hota: A higher order metric for evaluating multi-object tracking},
  author={Luiten, Jonathon and Osep, Aljosa and Dendorfer, Patrick and Torr, Philip and Geiger, Andreas and Leal-Taix{\'e}, Laura and Leibe, Bastian},
  journal={International journal of computer vision},
  volume={129},
  number={2},
  pages={548--578},
  year={2021},
  publisher={Springer}
}

@inproceedings{revach2021kalmannet,
  title={Kalmannet: Data-driven kalman filtering},
  author={Revach, Guy and Shlezinger, Nir and Van Sloun, Ruud JG and Eldar, Yonina C},
  booktitle={ICASSP 2021-2021 IEEE International Conference on Acoustics, Speech and Signal Processing (ICASSP)},
  pages={3905--3909},
  year={2021},
  organization={IEEE}
}

@article{revach2022kalmannet,
  title={KalmanNet: Neural network aided Kalman filtering for partially known dynamics},
  author={Revach, Guy and Shlezinger, Nir and Ni, Xiaoyong and Escoriza, Adria Lopez and Van Sloun, Ruud JG and Eldar, Yonina C},
  journal={IEEE Transactions on Signal Processing},
  volume={70},
  pages={1532--1547},
  year={2022},
  publisher={IEEE}
}

@article{choi2023split,
  title={Split-KalmanNet: A robust model-based deep learning approach for state estimation},
  author={Choi, Geon and Park, Jeonghun and Shlezinger, Nir and Eldar, Yonina C and Lee, Namyoon},
  journal={IEEE transactions on vehicular technology},
  volume={72},
  number={9},
  pages={12326--12331},
  year={2023},
  publisher={IEEE}
}

@article{buchnik2023latent,
  title={Latent-KalmanNet: Learned Kalman filtering for tracking from high-dimensional signals},
  author={Buchnik, Itay and Revach, Guy and Steger, Damiano and Van Sloun, Ruud JG and Routtenberg, Tirza and Shlezinger, Nir},
  journal={IEEE Transactions on Signal Processing},
  volume={72},
  pages={352--367},
  year={2023},
  publisher={IEEE}
}

@inproceedings{ni2024adaptive,
  title={Adaptive KalmanNet: Data-driven Kalman filter with fast adaptation},
  author={Ni, Xiaoyong and Revach, Guy and Shlezinger, Nir},
  booktitle={ICASSP 2024-2024 IEEE International Conference on Acoustics, Speech and Signal Processing (ICASSP)},
  pages={5970--5974},
  year={2024},
  organization={IEEE}
}

@article{buchnik2024gsp,
  title={GSP-KalmanNet: Tracking graph signals via neural-aided Kalman filtering},
  author={Buchnik, Itay and Sagi, Guy and Leinwand, Nimrod and Loya, Yuval and Shlezinger, Nir and Routtenberg, Tirza},
  journal={IEEE Transactions on Signal Processing},
  volume={72},
  pages={3700--3716},
  year={2024},
  publisher={IEEE}
}

@article{chen2025maml,
  title={MAML-KalmanNet: A neural network-assisted Kalman filter based on model-agnostic meta-learning},
  author={Chen, Shanli and Zheng, Yunfei and Lin, Dongyuan and Cai, Peng and Xiao, Yingying and Wang, Shiyuan},
  journal={IEEE Transactions on Signal Processing},
  year={2025},
  publisher={IEEE}
}

@article{dahan2025bayesian,
  title={Bayesian KalmanNet: quantifying uncertainty in deep learning augmented Kalman filter},
  author={Dahan, Yehonatan and Revach, Guy and Dunik, Jindrich and Shlezinger, Nir},
  journal={IEEE Transactions on Signal Processing},
  year={2025},
  publisher={IEEE}
}

@inproceedings{zhang2022bytetrack,
  title={Bytetrack: Multi-object tracking by associating every detection box},
  author={Zhang, Yifu and Sun, Peize and Jiang, Yi and Yu, Dongdong and Weng, Fucheng and Yuan, Zehuan and Luo, Ping and Liu, Wenyu and Wang, Xinggang},
  booktitle={European conference on computer vision},
  pages={1--21},
  year={2022},
  organization={Springer}
}

@article{aharon2022bot,
  title={BoT-SORT: Robust associations multi-pedestrian tracking},
  author={Aharon, Nir and Orfaig, Roy and Bobrovsky, Ben-Zion},
  journal={arXiv preprint arXiv:2206.14651},
  year={2022}
}

@article{du2023strongsort,
  title={Strongsort: Make deepsort great again},
  author={Du, Yunhao and Zhao, Zhicheng and Song, Yang and Zhao, Yanyun and Su, Fei and Gong, Tao and Meng, Hongying},
  journal={IEEE Transactions on Multimedia},
  volume={25},
  pages={8725--8737},
  year={2023},
  publisher={IEEE}
}

@inproceedings{cao2023observation,
  title={Observation-centric sort: Rethinking sort for robust multi-object tracking},
  author={Cao, Jinkun and Pang, Jiangmiao and Weng, Xinshuo and Khirodkar, Rawal and Kitani, Kris},
  booktitle={Proceedings of the IEEE/CVF conference on computer vision and pattern recognition},
  pages={9686--9696},
  year={2023}
}

@article{kalman1960new,
  title={A new approach to linear filtering and prediction problems},
  author={Kalman, Rudolph Emil},
  year={1960}
}

@book{bar2001estimation,
  title={Estimation with applications to tracking and navigation: theory algorithms and software},
  author={Bar-Shalom, Yaakov and Li, X Rong and Kirubarajan, Thiagalingam},
  year={2001},
  publisher={John Wiley \& Sons}
}

@article{tu2026tracking,
  title={Tracking and monitoring of pig behaviors using the RT-DETR+ BoT-SORT framework},
  author={Tu, Shuqin and Yang, Hairan and Mao, Liang and Tan, Baiyang and Zhang, Quan and Guo, Yunfeng and He, Ruilin},
  journal={Applied Animal Behaviour Science},
  volume={303},
  pages={107089},
  year={2026},
  publisher={Elsevier}
}

@article{guo2026enbot,
  title={EnBoT-SORT: Hierarchical fusion-association tracking with pseudo-sample generation for dense thermal infrared UAVs},
  author={Guo, Jinxin and Zhan, Weida and Chen, Yu and Zhu, Depeng and Jiang, Yichun and Xu, Xiaoyu and Han, Deng},
  journal={ISPRS Journal of Photogrammetry and Remote Sensing},
  volume={232},
  pages={138--154},
  year={2026},
  publisher={Elsevier}
}

@book{anderson2005optimal,
  title={Optimal filtering},
  author={Anderson, Brian DO and Moore, John B},
  year={2005},
  publisher={Courier Corporation}
}

@inproceedings{julier1997new,
  title={New extension of the Kalman filter to nonlinear systems},
  author={Julier, Simon J and Uhlmann, Jeffrey K},
  booktitle={Signal processing, sensor fusion, and target recognition VI},
  volume={3068},
  pages={182--193},
  year={1997},
  organization={Spie}
}

@inproceedings{bewley2016simple,
  title={Simple online and realtime tracking},
  author={Bewley, Alex and Ge, Zongyuan and Ott, Lionel and Ramos, Fabio and Upcroft, Ben},
  booktitle={2016 IEEE international conference on image processing (ICIP)},
  pages={3464--3468},
  year={2016},
  organization={Ieee}
}

@inproceedings{qin2025pptracker,
  title={PPTracker: Tracking UAV swarms with prior prompt},
  author={Qin, Haolin Qin and Li, Tianhao and Xu, Tingfa and Xu, Jingxuan and Fang, Yuqiang and Li, Jianan},
  booktitle={Proceedings of the Computer Vision and Pattern Recognition Conference},
  pages={6583--6590},
  year={2025}
}

@inproceedings{wang2025dist,
  title={Dist-tracker: A small object-aware detector and tracker for uav tracking},
  author={Wang, Wenzhen and Fu, Jing and Song, Jiayi and Li, Kaiyu and Qiao, Hui and Liu, Jiang and Sun, Hao and Cao, Xiangyong},
  booktitle={Proceedings of the Computer Vision and Pattern Recognition Conference},
  pages={6601--6609},
  year={2025}
}

@article{tian2026yolov12,
  title={Yolov12: Attention-centric real-time object detectors},
  author={Tian, Yunjie and Ye, Qixiang and Doermann, David},
  journal={Advances in neural information processing systems},
  volume={38},
  pages={78433--78457},
  year={2026}
}

@article{Jocher_Ultralytics_YOLO26_Unified_2026,
author = {Jocher, Glenn and Qiu, Jing and Liu, Mengyu and Lyu, Shuai and Akyon, Fatih Cagatay and Kalfaoglu, Muhammet Esat},
doi = {10.48550/arXiv.2606.03748},
title = {{Ultralytics YOLO26: Unified Real-Time End-to-End Vision Models}},
url = {https://arxiv.org/abs/2606.03748},
year = {2026}
}

@software{Robinson_RF-DETR_2025,
author = {Robinson, Isaac and Robicheaux, Peter and Popov, Matvei},
license = {Apache-2.0},
month = mar,
title = {{RF-DETR}},
url = {https://github.com/roboflow/rf-detr},
year = {2025}
}

@inproceedings{chen2023one,
  title={One-epoch training for object detection in fisheye images},
  author={Chen, Yu-Hsi},
  booktitle={Proceedings of the 5th ACM International Conference on Multimedia in Asia},
  pages={1--5},
  year={2023}
}

@inproceedings{fang2023depgraph,
  title={Depgraph: Towards any structural pruning},
  author={Fang, Gongfan and Ma, Xinyin and Song, Mingli and Mi, Michael Bi and Wang, Xinchao},
  booktitle={Proceedings of the IEEE/CVF conference on computer vision and pattern recognition},
  pages={16091--16101},
  year={2023}
}

@inproceedings{chen2024optimizing,
  title={Optimizing Facial Landmark Estimation for Embedded Systems Through Iterative Autolabeling and Model Pruning},
  author={Chen, Yu-Hsi and Tai, I-Hsuan},
  booktitle={2024 IEEE International Conference on Multimedia and Expo Workshops (ICMEW)},
  pages={1--6},
  year={2024},
  organization={IEEE}
}

@inproceedings{fang2024isomorphic,
  title={Isomorphic pruning for vision models},
  author={Fang, Gongfan and Ma, Xinyin and Mi, Michael Bi and Wang, Xinchao},
  booktitle={European Conference on Computer Vision},
  pages={232--250},
  year={2024},
  organization={Springer}
}

@article{meade2017spartan,
  title={Spartan HPC-cloud hybrid: delivering performance and flexibility},
  author={Meade, Bernard and Lafayette, Lev and Sauter, Greg and Tosello, Daniel},
  journal={University of Melbourne},
  volume={10},
  pages={49},
  year={2017}
}
}

\end{document}